\documentclass[
]{ceurart}

\usepackage{listings}
\usepackage{inputenc}
\usepackage{enumitem}
\usepackage{todonotes}

\begin{document}

\copyrightyear{2022}
\copyrightclause{Copyright for this paper by its authors.
  Use permitted under Creative Commons License Attribution 4.0
  International (CC BY 4.0).}

\conference{MediaEval'22: Multimedia Evaluation Workshop,
  January 13--15, 2023, Bergen, Norway and Online}

\title{Overview of The MediaEval 2022 Predicting Video Memorability Task}
\renewcommand{\shorttitle}{Predicting Video Memorability}
\renewcommand{\shortauthors}{ et al.}

\author[1]{Lorin Sweeney}
\author[2]{Mihai Gabriel Constantin}
\author[3]{Claire-Hélène Demarty}
\author[4]{Camilo Fosco}
\author[5]{Alba {G. Seco de Herrera}}[%
    % orcid=<orcid>,
    email=alba.garcia@essex.ac.uk,
    % url=<website>,
]\cormark[1]
\author[5]{Sebastian Halder}
\author[1]{Graham Healy}
\author[2]{Bogdan Ionescu}
\author[5]{Ana Matran-Fernandez}
\author[1]{Alan F. Smeaton}
\author[5]{Mushfika Sultana}

% Affiliations 
\address[1]{Dublin City University, Ireland}
\address[2]{University Politehnica of Bucharest, Romania}
\address[3]{InterDigital, France}
\address[4]{Massachusetts Institute of Technology Cambridge, USA}
\address[5]{University of Essex, UK}

\cortext[1]{Corresponding author.}

\begin{abstract}
This paper describes the 5th edition of the \textit{Predicting Video Memorability Task} as part of MediaEval2022. This year we have reorganised and simplified the task in order to lubricate a greater depth of inquiry. Similar to last year, two datasets are provided in order to facilitate generalisation, however, this year we have replaced the TRECVid2019 Video-to-Text dataset with the VideoMem dataset in order to remedy underlying data quality issues, and to prioritise short-term memorability prediction by elevating the Memento10k dataset as the primary dataset. Additionally, a fully fledged electroencephalography (EEG)-based prediction sub-task is introduced. In this paper, we outline the core facets of the task and its constituent sub-tasks; describing the datasets, evaluation metrics, and requirements for participant submissions.
\end{abstract}

\maketitle
%%%%%%%%%%%%%%%%%%%%%%%%%%%%
\section{Introduction}\label{sec:intro}
%%%%%%%%%%%%%%%%%%%%%%%%%%%%

As the natural world unwinds in an endless cacophony of sensory threads, the human brain selectively spins it into intelligible spools---filtering out information it deems unnecessary and spinning the rest into an intelligible internal representation. The human brain is an equally masterful weaver as it is spinster; weaving a colourful tapestry of meaning from its spools of intelligible threads by deciding which threads should be stitched into the canvas of our mind---what should be remembered and what should not.

The question is, what criteria does it use to decide what should and should not be remembered? Unfortunately, a satiating answer presently remains out of reach, leaving ``what it deems to be important'' as our appetizer. Memorability---the likelihood that a given piece of content will be recognised upon subsequent viewing---can accordingly be viewed as a proxy for human importance, which is what ultimately motivates and brings meaning to its exploration. After all, what could be more important than a measure of importance itself?

Memorability is accordingly the quintessential media metric by virtue of its proximal nature to the bedrock of human experience. If a system can predict the memorability of incoming information, it can evaluate its utility, then discard, filter, or augment the scantily useful, and ultimately curate more meaningful media content.

%%%%%%%%%%%%%%%%%%%%%%%%%%%%
\section{Related Work}\label{sec:work}
%%%%%%%%%%%%%%%%%%%%%%%%%%%%
The subject of memorability has seen an influx in interest since the likelihood of images being recognised upon subsequent viewing was found to be consistent across individuals~\cite{image2011memorable}. Driven primarily by the MediaEval Media Memorability tasks~\cite{ME2021,mediaeval2020memory,cohendet2019videomem, cohendet2018annotating}, recent research has extended beyond static images, pivoting to the more dynamic and multi-modal medium of video. In 2018, a video memorability annotation procedure was established, and the first ever large video memorability dataset VideoMem~\cite{cohendet2019videomem}---10,000 short soundless videos with both long-term and short-term memorability scores---was created. Additionally, the first ever analysis of human consistency and video memorability was conducted. In 2019, the task ran for a second time using the same dataset allowing participants to learn from the previous year's task and carry out comparative analysis of results from one year to the next. In 2020, a new smaller dataset was introduced which included audio for the first time~\cite{kiziltepe2021annotated}. In 2021, that dataset was extended, with a second large short-term dataset---Memento10k~\cite{mem10k}---being released, short-term memorability was sub-categorised into \textit{raw} and \textit{normalised} scores, an optional generalisation sub-task was proposed, and a pilot EEG study~\cite{sweeney2021overview} was conducted. 

Over the course of those four tasks, we have learned that short-term video memorability is easier to predict than long-term memorability, simple image features, such as hue, saturation, or spatial frequency, have repeatedly been found not to correlate with memorability, properties such as aesthetics and interestingness likewise do not correlate with memorability, ensembles that combine different modalities provide the best results, combining deep visual features in conjunction with semantically rich features such as captions, emotions, or actions~\cite{azcona2019ensemble,mem10k,sweeney2021influence, zhao2020multi} is a highly effective approach, dimensionality reduction improves prediction results, and certain semantic categories of objects or places are more memorable than others~\cite{cohendet2019videomem, mem10k}.

%%%%%%%%%%%%%%%%%%%%%%%%%%%%
\section{Description}\label{sec:description}
%%%%%%%%%%%%%%%%%%%%%%%%%%%%
In this edition, the \textit{Predicting Video Memorability} task challenges participants to develop systems that automatically predict short-term memorability scores for short form videos. Participants are provided with three datasets and offered three sub-tasks in which to participate.

\subsection{Sub-task 1: How memorable is this video? - Video-based prediction}

Using the Memento10k~\cite{mem10k} dataset, participants are required to generate automatic systems that predict short-term memorability scores of new videos based on the given video dataset and their memorability scores.

\subsection{Sub-task 2: How memorable is this video? - Generalisation (optional)}

Sub-task 2 is a natural extension of sub-task 1, where participants can evaluate their systems from sub-task 1 (trained on Memento10k) on the VideoMem dataset.
Alternatively, participants can also train a system on the VideoMem dataset and evaluate it on Memento10k.

\subsection{Sub-task 3: Will this person remember this video? - EEG-based prediction (optional)}

Participants are required to generate automatic systems that predict whether or not a given subject will recognise a given video upon subsequent viewing (N.B., this differs from memorability as it is subject specific and a binary prediction, rather than subject agnostic and a floating point prediction) based on the provided EEG data. Participants may choose to use the provided EEG features in concert with sub-task 1's visual features or in isolation. However, they must use the EEG features in some capacity.

%%%%%%%%%%%%%%%%%%%%%%%%%%%%
\section{Dataset Details}
\label{sec:details}
%%%%%%%%%%%%%%%%%%%%%%%%%%%%
%
In the interest of clarity, standardisation, and the facilitation of more directed inquiry, we have narrowed the scope of the tasks forgoing with raw and long-term memorability scores in favour of normalised short-term scores. Additionally, in order to address systemic data quality issues highlighted by a consistent disparity between participant systems trained on the TRECVid2019 dataset and the Memento10k dataset, we have opted to replace the TRECVid2019 dataset with VideoMem, and to elevate Memento10k to primary dataset status. Additionally, a fledged EEG dataset (EEGMem) is provided.

The following set of pre-extracted features are provided along with the Memento10k and VideoMem datasets: 

\begin{itemize}[leftmargin=*]
  \item Image-level features: AlexNetFC7~\cite{krizhevsky2017imagenet}, HOG~\cite{dalal2005histograms}, HSVHist, RGBHist, LBP~\cite{he1990texture}, VGGFC7~\cite{simonyan2014very}, DenseNet121~\cite{huang2017densely}, ResNet50~\cite{he2016deep}, EfficientNetB3~\cite{tan2019efficientnet}
  \item Video-level features: C3D~\cite{tran2015learning}
\end{itemize}

\noindent 
Three frames---the first, middle, and last---from each video were used to extract image-level features.

\subsection{Memento10k}
Memorability scores were collected through \textit{Memento: The Video Memory Game}, a memorability experiment predicated the old-new recognition paradigm~\cite{image2011memorable}, where
crowdworkers from Amazon’s Mechanical Turk (AMT) watch a continuous stream of three-second video clips, and are asked to press the space bar when they see a repeated video. To maximise the pace and keep the experiment engaging, videos are shown as a continuous stream. When participants press their spacebar, they receive either a red (incorrect) or green (correct) flash as feedback. If a repeat is correctly identified, known as a ``hit”, the stream skips ahead to the next video; there is no feedback for missed repeats. Each level of the memory game contains on average 204 videos (with repeats) and lasts $\sim$ 9 minutes. The number of intervening videos between the first and second occurrence of a repeated video is known as the ``lag”. The game consists of “vigilance” repeats that occur at short lags of 2-3 videos and are used to filter out inattentive workers and “target” repeats at lags of 9-200 videos that provide memorability data.

The Memento10k dataset~\cite{mem10k} consists of 10,000 three-second videos depicting in-the-wild scenes, each with associated short-term memorability scores, memorability decay values, action labels, and five human generated captions. The memorability scores were computed with an average of 90 annotations per video, and the videos were deafened before being shown to participants. 7,000 videos are released as part of the training set, and 1,500 are provided for validation. The remaining 1,500 videos are kept for the official test set.

\subsection{VideoMem}
The VideoMem dataset~\cite{cohendet2019videomem} consists of 10,000 soundless seven-second videos each with associated short-term and long-term memorability scores, however, long-term scores are omitted from this year's task. Videos were extracted from cinematic raw stock footage and come with a caption. 7,000 videos are released as part of the training set, and 1,500 are provided for validation. The remaining 1,500 videos are kept for the official test set.

\subsection{EEGMem}
%
%EEGMem is not a standalone dataset, but rather, a supplemental set of annotations for Memento10k~\cite{mem10k} which consists of pre-extracted features from EEG recordings of subjects while watching a subset of the videos. Pre-extracted features include Event-Related Potentials (ERPs), Event-Related Spectral Perturbations (ERSPs), and spectrograms. 
The EEGMem dataset comprises pre-extracted features from EEG recordings for 12 subjects captured while they watched a subset of the Memento10k~\cite{mem10k} videos. Participants watched the same videos again through a custom-built online portal between 24--72~hours after the video-EEG recording session, where they were required to indicate for each video whether or not they recognised it, providing binary ground truth annotations\footnote{Further details on the EEGMem dataset and data collection protocol are available at: https://bit.ly/3BTstj7 }.

%%%%%%%%%%%%%%%%%%%%%%%%%%%%
\section{Evaluation}
%%%%%%%%%%%%%%%%%%%%%%%%%%%%
%
A total of five runs can be submitted by each participant for each sub-task. For sub-task 1 all information relating to the Memento10k dataset, i.e., ground-truth data, annotation data, pre-extracted features, and features extracted from provided material, may be used to build the system. For sub-task 2, in similar fashion to sub-task 1, all information relating to the Memento10k and VideoMem datasets may be used to build the system, however, only one dataset may be used per run, and must be evaluated on the other dataset to assess generalisability. For sub-task 3 the only requirement is that EEG data must be, to some extent, included in the system.

Three standard metrics will be used to assess participant system performance for sub-tasks 1 and 2: Spearman’s rank correlation, Pearson correlation, and mean squared error. However, similar to previous years, Spearman’s rank correlation will be adopted as the official metric as it enables inter-method comparisons by taking into account monotonic relationships between ground-truth data and system output. Submissions for sub-task 3 will be evaluated using the Area Under the Receiver Operating Characteristic Curve.

%%%%%%%%%%%%%%%%%%%%%%%%%%%%
\section{Conclusions}
%%%%%%%%%%%%%%%%%%%%%%%%%%%%
%
This paper presents an overview of the fith edition of the MediaEval Predicting Video Memorability task. Similar to previous years, the task presents a framework to evaluate the prediction of the memorability of short form videos. This year the task focuses on short-term memorability and introduces a task based on EEG signals.
Details regarding the participants’ approaches and their results
can be found in the proceedings of the 2022 MediaEval workshop\footnote{See CEUR Workshop Proceedings (\url{CEUR-WS.org}).}.
%%%%%%%%%%%%%%%%%%%%%%%%%%%%
\section*{Acknowledgements}
%%%%%%%%%%%%%%%%%%%%%%%%%%%%
Science Foundation Ireland under Grant Number SFI/12/RC/2289\_P2, cofunded by the European Regional Development Fund.
Financial support also provided by the University of Essex Faculty of Science and Health Research Innovation and Support Fund.
Financial support also provided under project AI4Media, a European Excellence Centre for Media, Society and Democracy, H2020 ICT-48-2020, grant \#951911.

\def\bibfont{\small} % comment this line for a smaller fontsize
\bibliography{references} 

\end{document}